\documentclass{article}

\usepackage{microtype}
\usepackage{graphicx}
\usepackage{subfigure}
\usepackage{booktabs} 

\usepackage{hyperref}

\usepackage{kpfonts}

 \newcommand{\add}[2][]{\todo[linecolor=blue,backgroundcolor=blue!25,bordercolor=blue,#1]{#2}}
 \newcommand{\Add}[2][]{\todo[inline,linecolor=blue,backgroundcolor=blue!25,bordercolor=blue,#1]{#2}}

\newcommand{\OUR} {{\sc BounceGrad}}

\usepackage{amsmath,amssymb}

\usepackage{amsmath,amsfonts,bm}

\def\eqref#1{equation~\ref{#1}}

\def\1{\bm{1}}

\def\va{{\bm{a}}}
\def\vb{{\bm{b}}}

\def\vd{{\bm{d}}}
\def\ve{{\bm{e}}}
\def\vf{{\bm{f}}}
\def\vg{{\bm{g}}}

\def\vn{{\bm{n}}}

\def\vr{{\bm{r}}}

\def\vv{{\bm{v}}}

\def\vx{{\bm{x}}}

\DeclareMathAlphabet{\mathsfit}{\encodingdefault}{\sfdefault}{m}{sl}
\SetMathAlphabet{\mathsfit}{bold}{\encodingdefault}{\sfdefault}{bx}{n}

\def\sF{{\mathbb{F}}}
\def\sG{{\mathbb{G}}}

\def\sI{{\mathbb{I}}}

\def\sL{{\mathbb{L}}}

\def\sO{{\mathbb{O}}}

\def\sX{{\mathbb{X}}}

\def\sZ{{\mathbb{Z}}}

\DeclareMathOperator*{\argmin}{arg\,min}

\usepackage{bbm}
\usepackage{enumitem}
\setitemize{noitemsep,topsep=0pt,parsep=0pt,partopsep=0pt}

\usepackage[accepted]{icml2019} 

\icmltitlerunning{Graph Element Networks}
\begin{document}

\twocolumn[
\icmltitle{Graph Element Networks: adaptive, structured computation and memory}



\begin{icmlauthorlist}
\icmlauthor{Ferran Alet}{csail}
\icmlauthor{Adarsh K. Jeewajee}{csail}
\icmlauthor{Maria Bauza}{meche}\\
\icmlauthor{Alberto Rodriguez}{meche}
\icmlauthor{Tom\'{a}s Lozano-P\'{e}rez}{csail}
\icmlauthor{Leslie Pack Kaelbling}{csail}
\end{icmlauthorlist}

\icmlaffiliation{csail}{CSAIL - MIT, Cambridge, MA, USA}
\icmlaffiliation{meche}{Mechanical Engineering - MIT, Cambridge, MA, USA}

\icmlcorrespondingauthor{Ferran Alet}{alet@mit.edu}

\icmlkeywords{Graph Neural Networks, Memory, Physics, Robotics}

\vskip 0.3in
]



\printAffiliationsAndNotice{}  
\begin{abstract}

We explore the use of graph neural networks (GNNs) to model spatial processes in which there is no {\em a priori} graphical structure.  Similar to {\em finite element analysis}, we assign nodes of a GNN to spatial locations and use a computational process defined on the graph to model the relationship between an initial function defined over a space and a resulting function in the same space.   We use GNNs as a computational substrate, and show that the locations of the nodes in space as well as their connectivity can be optimized to focus on the most complex parts of the space.  Moreover, this representational strategy allows the learned input-output relationship to generalize over the size of the underlying space and run the same model at different levels of precision, trading computation for accuracy.  We demonstrate this method on a traditional PDE problem, a physical prediction problem from robotics, and learning to predict scene images from novel viewpoints.

\end{abstract}
\section{Introduction}
\label{sec:introduction}
A great deal of success in deep learning has come from finding appropriate structural inductive biases to impose on network architectures. For an architectural assumption to be useful, it has to exploit a structural property that is (approximately) satisfied in a broad set of tasks. For instance, convolutional neural networks exploit the locality and spatial invariance found in many computer vision problems. Similarly, graph neural networks (GNNs) exploit underlying relational structure,  which makes them a good fit for tasks consisting of a set of entities with pairwise interactions.

Traditional applications of GNNs assume an {\em a priori} notion of entity (such as bodies, links or particles) and match every node in the graph to an entity.  We propose to apply GNNs to the problem of modeling transformations of functions defined on continuous spaces, using a structure we call \textit{graph element networks} (GENs). Inspired by finite element methods, we use graph neural networks to mesh a continuous space and define an iterative computation that propagates information from some sampled input values in the space to an output function defined everywhere in the space.
GENs allow us to model systems that have spatial structure but lack a clear notion of entity, such as the spatial temperature function in a room or the dynamics of an arbitrarily-shaped object being pushed by a robot.

\begin{figure*}
    \centering
    \includegraphics[width=\linewidth]{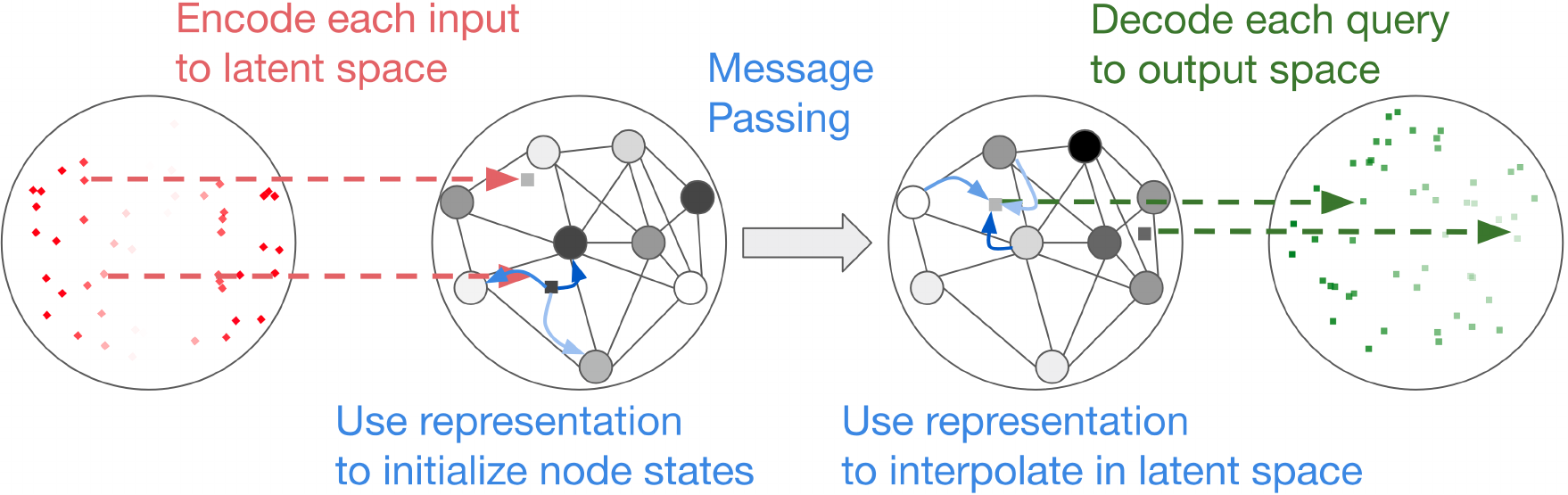}
    \vspace{-0mm}
    \caption{Graph element networks (GENs) map input spatial functions to output spatial functions over a metric space $\sX$ (in this figure, a disk). Learned encoders embed samples from the input to a latent space and a representation function allows us to map these embeddings to the initial state of a graph neural network (GNN). We then run the GNN for several steps to propagate information between the nodes, which makes the local information global.  Finally, for a given query, we interpolate the latent state of any point using a linear weighting of the latent states of a subset of nodes and use a learned decoder to make our prediction.}
    \label{fig:GEN_diagram}
    \vspace{-0mm}
\end{figure*}

Finite element methods~\cite{hughes2012finite} are used to numerically solve  partial differential equations (PDEs) by dividing a space into small subparts (elements) where the PDE can be approximated. Analogously, in a GEN, we place nodes of a GNN in the space, allowing each node to model the local state of the system, and establish a connectivity graph among the nodes (it may be a regular mesh or any other arbitrary graph).  Then, input values are specified for multiple spatial locations, and used to initialize the states of the GNN nodes, which are then propagated according to the GNN's update functions.
Finally, output values may be read back from any point in the system, by interpolating between values of nodes in the GNN.  A critical aspect of the GEN is that although the model has a fixed set of weights, it can be used to model small or large spaces, with coarse, fine or variable-resolution sampling of the space by reusing the same weights in different nodes and edges.


Although GENs were originally inspired by finite element methods that model PDEs, we show that they are much more widely applicable.  We are able to model pushing of arbitrarily shaped objects, whose discrete dynamics do not follow a PDE, with the same structure. Furthermore, we apply GENs to the task of scene representation, where the system is presented with  images of a scene from various camera poses and is asked to predict the view from an unseen pose.  We show that GENs naturally adapt to scenes of increasing size without further training.


In the following sections, we discuss related work, define the graph element network model and provide an algorithm for training it from data.  We conclude with experimental results.

\section{Background and related work}

Our method builds directly on graph neural networks.  We describe that class of models here, and outline particularly relevant work in using neural networks to model continuous-space systems.

\subsection{Graph neural networks} \label{subsec:GNN}

Graph neural networks~\citep{sperduti1997supervised,gori2005new,scarselli2009graph} 
(see recent surveys of~\citet{battaglia2018relational,zhou2018graph,wu2019comprehensive}) perform computations over a graph, with the aim of incorporating a \textit{relational inductive bias}:  assuming the existence of a set of entities and relations between them. In general, GNNs are applied in settings where nodes and edges are explicit and match concrete entities of the domain; for instance, nodes can be objects~\citep{chang2016compositional,battaglia2016interaction,hamrick2018relational} or be parts of objects~\citep{wang2018nervenet,mrowca2018flexible,li2018learning}. GNNs are typically trained with supervised values for each node.
Of note is the work of~\citet{van2018relational}, which assumes a notion of entity but learns what entities are without supervision. 

We build on GNNs, but do not assume that the domain has a prior notion of entity.
{\em Non-local neural networks}~\cite{wang2018non} also use GNNs to model long-range interactions in video without a traditional notion of entity, but the nodes are tied to a CNN channel and the graph is required to be complete, which limits the applicability of the model. 
The field of {\em geometric deep-learning} \cite{Bronstein2017GeometricDL} aims to build deep networks for graphs \cite{bruna2013spectral,defferrard2016convolutional,kipf2016semi} and non-Euclidean manifolds~\cite{maron2017convolutional}. Most similar to our work,~\citet{monti2017geometric} build fixed graphs of super-pixels from the input and~\citet{smith2019geometrics} propose to output refinable meshes as 3D models for objects. In contrast, we propose to use GNNs as an optimizable computational substrate, rather than an object of interest per se. Finally, we believe ideas designed for manifolds (such as~\citet{jaderberg2015spatial}) and graphs (such as~\citet{velickovic2017graph}) can help improve our representation functions and graph propagation phases, respectively.


We build on the framework of {\em message-passing neural networks} (MPNNs)~\cite{gilmer2017neural}, similar to {\em graph convolutional networks}(GCNs)~\citep{kipf2016semi,battaglia2018relational}. In MPNNs, each node and each edge have a corresponding neural network module, which is usually a small feed-forward network. The same module can be reused across many nodes or edges, a similar structure to convolutional networks, which reuse the same filter across all positions in the image. 
MPNNs store a hidden state value at every node in the graph.  An MPNN is specified by $\langle N, E, m_e, m_v\rangle$, where $N$ is a finite set of  nodes $v_1,\dots, v_r$, $E$ is a set of directed edges between pairs of nodes, $m_e$ is an {\it edge module} neural network mapping the hidden values of the nodes it connects into a {\em message} to the target node of the edge, and $m_v$ is a {\em node module} neural network mapping the current value of a node and the aggregated incoming messages into a new node value.  The input to an MPNN is the initial hidden state of every node in the graph: $h^0_1, \dots, h^0_r$. From this, we apply $T$ steps of neural message passing to obtain the final node states: $h^T_1, \dots, h^T_r$. In each step of message passing: 
(1) Each edge $(n_i, n_j)$ uses the edge module $m_e$ to compute its \textit{message} $m^{t+1}_{ij} = m_e\left(h^t_i,h^t_j\right)$; 
(2) Each node aggregates all its incoming messages into its pre-update value  $u^{t+1}_j = \sum_{(i, j) \in E} m^{t+1}_{ij}$; and 
(3) Each node updates its hidden state from its pre-update value using the node module: $h^{t+1}_j =  m_v\left(h^t_j, u^{t+1}_j\right)$.
This computation has the same structure as a message-passing algorithm in a Markov random field, but the messages in an MPNN need not have a semantic interpretation in terms of probability.

\subsection{Applications of GENs and related models}
There is existing work that casts finite element methods as feed-forward neural networks~\cite{takeuchi1994neural,Ramuhalli2005FiniteelementNN} or Hopfield networks~\cite{Guo1999FiniteEA,al2008solutions}  as well as mesh-free methods to solve PDEs~\cite{dissanayake1994neural,vanMilligen1995NeuralND,leake2018deep,Long2018PDENetLP,sirignano2018deep}. Others, such as~\citet{james2018machine} have used data from a PDE solver to train a neural network.  In contrast, rather than learning to map an input directly to the global solution, our approach innovates by learning a simpler local model and then using message passing to derive the global solution.  The  GEN approach enables better generalization over domain shapes and sizes, and provides a much stronger inductive bias;  it also allows us to address a variety of other learning problems, as illustrated in our experiments section.  

Similar to GENs, neural ordinary differential equations~\cite{chen2018neural} can trade off accuracy with computation, while keeping the model parameters constant. However, they address a substantially different problem: they can learn continuous-time ordinary differential equations whose dynamics are parametrized by a neural network whereas we consider continuous-space partial differential equations.  

Image-to-image problems~\cite{isola2017image} are a subcase of spatial function transformation problems(section \ref{subsec:SFT}). In these problems, GENs can be substitutes for fully convolutional networks~\cite{long2015fully} and spatial transformer networks~\cite{jaderberg2015spatial}, being more complex but also more flexible and expressive.

Finally, there are interesting connections to be made between GENs and grid cells~\cite{hafting2005microstructure}, which help animals create spatial representations by meshing the space in hexagonal grid patterns. Of note are two concurrent discoveries~\cite{boccara2019entorhinal,butler2019remembered} that suggest these grid patterns warp to focus more on interesting parts of the space, such as places where an animal found reward; mechanisms similar to the mesh optimization in section~\ref{subsec:optimize} might account for this phenomenon. 

\subsection{Mapping a function to itself} \label{subsec:meta-learning}
We can also see GENs as performing a form of meta-learning, in the case that the input and output functions are the same.  Here, the fully-trained GEN model takes in a training data set of $(x, f(x))$ pairs as input and can predict values $\widehat{f}(x)$ at any new query point in the space.  The underlying learning algorithm is a form of exemplar-based regression, with the meta-learning process adapting the parameters of the process used to encode the training data into the exemplar values at the node points and to decode those values to answer queries.  Graph neural networks have been used previously for meta-learning by~\citet{garcia2017few}, similar to work by~\citet{choma2018graph}, by allocating a node to each training and testing example and training the GNN to perform inference in the form of label propagation. Concurrent work by~\citet{kim2019attentive} introduces
attentive neural processes, which build upon Neural Processes~\cite{garnelo2018neural} by adding attention mechanisms between inputs and from queries to inputs, thus also performing a propagation in the fully-connected graph of inputs. In contrast, apart from applying it to function transformation problems, we can have many fewer nodes than training examples and sparse connectivity, making it computationally more efficient, and (in contrast to~\citet{garcia2017few}) do not need to know the test points {\it a priori}. Moreover, if inputs and queries only cover some parts of the space, the topology of the graph on examples may not be that of the underlying space. For example, in our weather model example of section~\ref{subsec:SFT}, we may have data from many weather sensors in cities but none in oceans, thus not capturing vast parts of the Earth that have a significant influence on weather dynamics. 

GENs can also be seen as a type of memory system, writing (inputs) into and reading (queries) from a mesh that organizes memories in space, allows them to influence one another via neighborhood relations, and integrates them locally to answer queries.  They have the advantages of flexible size and a structure that mimics that of the underlying space. In that regard, they are related to other external memory systems that support read and write operations via attention~\cite{Ma2018ATF,graves2014neural,kurach2015neural,graves2016hybrid,Glehre2018DynamicNT} and especially to neural SLAM and neural map~\cite{Zhang2017NeuralS,Parisotto2017NeuralMS}, which are also spatially grounded. Notably, \citet{Fraccaro2018GenerativeTM} used differentiable neural dictionaries (DNDs)~\cite{Pritzel2017NeuralEC} to solve a problem similar to our scene representation experiment in section~\ref{subsec:GQN_exp}. Compared to many of these methods, GENs are not restricted to Euclidean spaces, have a cost independent of the number of memories and are optimizable because nodes need not be in particular positions in space. Moreover, they are active; i.e., the mesh performs computations that combine memories to integrate local information with global context.
\section{Graph element networks}
We first describe the class of {\em spatial function transformation} (SFT) problems and the class of {\em graph element network} (GEN) models, and show how to train the parameters of a GEN given data from an SFT problem. 

\subsection{Spatial function transformation problems}\label{subsec:SFT}
Let $\sX$ be a bounded metric space, let $\sI_1,\dots,\sI_c$ and $\sO_1, \dots, \sO_{c'}$ be (not necessarily bounded) metric spaces. Let $f_1,\dots,f_c,g_1,\dots,g_{c'}$ be functions s.t. $f_i: \sX \rightarrow \sI_i$ and $g_j: \sX \rightarrow \sO_j$.

A {\em spatial function transformation} (SFT) is a mapping from the sequence of functions $\vf=(f_1,\dots,f_c)\in\sF$ to the sequence of functions $\vg=(g_1,\dots,g_{c'})\in \sG$. 
We describe both the input functions $\vf$ and output functions $\vg$ with function samples.  The input function is specified by a set of data points $(x\in \sX, i \in (1..c), s\in \sI_i)$ that specify the value $s$ for input dimension $i$ at location $x$.  The output function is queried at a set of locations and dimensions $(x\in \sX, j \in (1..c'))$, resulting in output values $s\in \sO_j$.

As an example, consider learning a simplified weather model that predicts an output function (the amount of rain tomorrow at any point on Earth) given some input functions (the current relative humidity and wind velocity). Then, $\sX=S^2$ (Earth's surface), $\sI_1=[0,1]$ (relative humidity), $\sI_2=\mathbb{R}^3$ (wind velocity) and $\sO_1=\mathbb{R}^+$ (amount of rain tomorrow). Given training pairs of input and output functions, we learn how to predict the amount of rain tomorrow at a set of query positions, given some humidity and wind measurements (which need not be in the same positions as the queries).

Another familiar class of examples are image-to-image problems such as semantic segmentation, where the space is the image canvas, the input is in color space ($\sI_1=[0,255]^3$) and the output in semantic space ($\sO_1=1_\kappa$, the probability simplex of the semantic classes).

\subsection{Learning an SFT}
Our objective is to learn SFTs from data, which consists of
many $\left(\vf,\vg\right)$ pairs. For each pair, we have multiple function evaluations which serve as input samples (if they are from a function $f_i$) or as query and target (if they are from a function $g_j$). More concretely, our data is a set of tuples of sets of evaluations 
$\left\{\left\langle \mathcal{D}^\text{inp}_{p},\mathcal{D}^\text{out}_{p}\right\rangle_{p=1}^P\right\}$
where 
$\mathcal{D}^\text{inp}_{p}\in\prod_{l_1=1}^{L_1} \left(\sX\times (1..c) \times \sI_{i_{l_1}}\right)$ and $\mathcal{D}^\text{out}_{p}\in\prod_{l_2=1}^{L_2}\left(\sX\times (1..c') \times \sO_{j_{l_2}}\right)$. We have omitted the dependence of ${L_1},L_2$ (input and output set sizes) and $i_{l_1},j_{l_2}$(input and output space indices) on function pair $p$ to simplify notation. 

Our hypothesis space $\mathcal{H}$ is a set of
functions mapping from functions $\vf$ to functions $\vg$; so each $h \in \mathcal{H}$ maps $\sF\rightarrow\sG$.
Naturally, our loss depends on the error made between our prediction $h\left(\vf\right)$ and the true function $\vg$. Therefore
\begin{align*}
\mathcal{L}(h) &= \frac{1}{Pc'} \sum_{p=1}^P\sum_{j=1}^{c'} \mathcal{L}_j\left(h(\vf^p)_j,\vg_j^p\right) \\
&\approx \frac{1}{PL_2} \sum_{p=1}^P \sum_{l=1}^{L_2} \mathcal{L}_{j_{l}}\left(h(\mathcal{D}^\text{inp}_{p})_{j_{l}}(x_{j_{l}}),\vg^p_{j_{l}}(x_{j_{l}})\right)
\triangleq\widehat{\mathcal{L}}(h)
\end{align*}
where we have approximated the loss in function space by the expected loss of the samples and approximated the function $h(\vf)$ by $h$ (abusing notation) of the input samples from $\vf$.  We let the loss of each sample in space $\sO_{j_{l}}$, $\mathcal{L}_{j_{l}}$, simply be the distance in that metric space. Our learning problem is then to find $h^* = \argmin_{h\in \mathcal{H}} \widehat{\mathcal{L}}(h)$.
\subsection{Graph element networks}
A {\em graph element network} (GEN) is a tuple $\langle \sX, \sF, \sG, \sL, N, E, \ve, \vd, m_e, m_v, T \rangle$, where
\begin{itemize}
    \item $\sX$, $\sF$, and $\sG$ are the spaces of an SFT;
    \item $\sL$ is a {\em latent} metric space; 
    \item $N$ is a finite set of $n$ {\em nodes}, where each node $v_l$ is placed at location $x_l \in \sX$;
    \item $E \subseteq N \times N$ is a finite set of {\em edges} between nodes;
    \item $\ve = e_1, \ldots, e_c$ is a sequence of {\em encoders}, where $e_i$ is  a mapping from $\sI_i \rightarrow \sL$, represented as a feed-forward neural network with weights $W_e^i$;
    \item $\vd = d_1, \ldots, d_{c'}$ is a sequence of {\em decoders}, where $d_j$ is a mapping from $\sL \rightarrow \sO_j$, represented as feed-forward neural network with weights $W_d^j$;
    \item $m_e$ and $m_v$ are the edge and vertex modules of a GNN defined on the latent space $\sL$;
    \item $T$ is the number of rounds of message-passing done by the GNN.
\end{itemize}
The set of input sample values $(x, i, s)$ is encoded into initial values $z_v^0 \in \sL$ for each node $v$.  Then, $T$ rounds of message passing take place in the GNN defined by $E$, $m_e$, and $m_v$, resulting in values $z_v^T$ at each node.  Finally, these values can be decoded into a value for any output function at any query point. See figure~\ref{fig:GEN_diagram} for an illustration.

Both the encoding and decoding processes require a strategy for assigning values to all points in $\sX$ given values at a finite subset of points.  To do this, we require a {\em representation function} $r : \sX \rightarrow \mathbb{R}^n$, which maps any point in $\sX$ to a vector of "coordinates" or "weights" on the nodes.  Given $r$, 
we compute the initial latent value at each node as a weighted sum of the encoded input values, 
$z_l^0 = \sum_{(x, i, s) \in \mathcal D^\text{inp}} r(x)_l e_i(s)$.  
After $T$ rounds of message passing, we have values $z_l^T$ at each node.  
We can extrapolate from these latent values at the nodes to a latent function over the space $\sX$ as $z(x) = \sum_l r(x)_l z_l^T$.  Finally, we define the output function $\widehat{g}(x) = d(z(x))$ on all dimensions, by applying the decoder to the latent function value $z(x)$. Note that we make the assumption that the latent space is locally linear, which does not imply that either input or output functions are locally linear.

In summary, first, the GEN uses an encoder to go from input space to latent space, the representation function to locally interpolate the initial node states and the GNN to make the local information global. Then, symmetrically, we use the same representation function to locally interpolate the final latent vector at the query positions and the decoder to map from latent space to output space. The GEN is parameterized by the weights of the encoder and decoder networks as well as those of the edge and vertex modules of the GNN.  These weights can be trained to minimize $\widehat{\mathcal L}$ end-to-end via gradient descent. For more information, please refer to the appendix and the code.

\subsection{Optimizing graph topology} \label{subsec:optimize}
Thus far, we have assumed that the number, connectivity, and positions of the nodes in the GEN are fixed, but in fact they can also be adjusted during training.  
First, since nodes are not tied to any entities and GNNs can have any number of nodes, we can train \textit{the same} network on multiple mesh sizes, being able to choose the desired mesh size at test time to trade off accuracy with compute (section \ref{subseq:Poisson_exp}) or adapt to different problem sizes(section \ref{subsec:GQN_exp}).

Second, we note that the output of the GEN is differentiable with respect to the positions of the nodes, so they can be easily adjusted via gradient descent, with the effect of allocating representational capacity to the parts of the space where it is most useful.
We often use a regular pattern of graph connectivity, such as a Delaunay triangulation, which automatically computes the edge set from the positions of the nodes. If we adjust the nodes' positions, then, it is straightforward to dynamically compute the graph edges after each update to the node positions. Changing connectivity in this manner changes the form of the computation done by the GNN in a way that is non-smooth, but we have not observed convergence problems from doing so. Experiments illustrating this idea can be found at the end of section \ref{subseq:Poisson_exp}.

It is possible to combine gradient-descent with discrete non-local search steps in our optimization process that add or remove nodes from the network, in the style of the {\sc BounceGrad} algorithm~\cite{alet2018modular}.  This process would allow the network to grow to a size appropriate to a given problem instance, but we have not yet explored it experimentally.

\section{Experiments}
We do a varied set of experiments to illustrate the diversity of applications of GENs. First, we learn a model and a solution process that solves a partial differential equation. Then, we show that GENs can be used to predict the result of physical interactions of arbitrarily shaped objects.  
Finally, we apply GENs to the scene representation problem introduced by~\cite{eslami2018neural} and show that GENs can outperform generative query networks (GQNs) by generalizing to very different input scales at runtime.

We use neural processes (NPs)~\cite{garnelo2018neural}, which were also used by~\citet{eslami2018neural}, as the baseline for comparison in all our scene representation experiments. 
Neural processes take $(x_i,y_i)$ pairs; encode them using neural encoder $e$; aggregate those outputs (in a permutation-invariant way such as summation) into a common representation $r$; concatenate the query point $x_q$; and finally feed the concatenation to a decoder $d(x_q,r)$. Neural processes also have a customization that allows them to give meaningful uncertainty estimates, but we do not use it since it is not relevant for our experiments. Although the original NP work considers a more limited setting (with a single space $\sI_1=\sO_1$), they use a compatible input-output specification and can handle inputs in any metric space $\sX$. We can see an NP as a simple GEN with a single node and no message passing steps. By giving more capacity to the encoder and decoder we will thus check that adding the GEN's graph structure is useful.

For brevity, minor experimental details are in the appendix. Code can be found at \url{https://github.com/FerranAlet/graph_element_networks}.

\subsection{Modeling partial differential equations} \label{subseq:Poisson_exp}
Finite element methods (FEMs) are typically used to find numerical solutions to partial differential equations (PDEs) and share with GENs the form of a local computational structure over nodes embedded in $\sX$, so PDEs are a natural first class of problems to try.  PDEs are equations characterizing the partial derivatives of a function with respect to its spatial coordinates. In this example, we use GENs to model the Poisson equation on different manifolds.  Crucially, 
our system does not know the analytical form of the equation, but learns it from input-output training pairs of functions so that, given data describing the initial state of a new system, it can produce a solution.

The \textit{Poisson equation} is the best-known PDE, used to model gravitational or electromagnetic fields and also the heat equation. The solution is a function $\phi(x,y)$ that satisfies 
$$\Delta \phi = \nabla^2 \phi = \left(\frac{\partial^2}{\partial x^2} + \frac{\partial^2}{\partial y^2}\right) \phi(x,y) = \psi(x,y)\;\;.$$
The function $\psi$ defines sources and sinks, and the solution may be constrained by boundary conditions that specify some values or constraints on values of $\phi$.

\paragraph{Solving a PDE on a square} In the first experiment, we use a GEN to learn to model the propagation of heat in two-dimensional "houses", each with a different set of heaters(sources), coolers(sinks) and exterior temperature values(boundary conditions): \begin{itemize}
    \item Any point inside the house, but outside a heater follows the Laplace equation, i.e., $\psi(x,y)=0$;
    \item Any point in a heater or cooler follows the Poisson equation with $\psi(x,y)=C_h$, a known constant;
    \item All exterior borders are modeled as very thin windows, thus being at the exterior temperature $T$, a known constant, which imposes \textit{Dirichlet boundary conditions}; i.e. fixes $\phi(x,y)=T$ for all $(x,y)$ in an exterior wall.
\end{itemize}
For training, we get ground-truth solutions using a FEM with a very dense mesh of $250^2$ nodes, computed with the program FEniCS~\cite{alnaes2015fenics}. 

\begin{figure}
    \centering
    \includegraphics[width=0.75\linewidth]{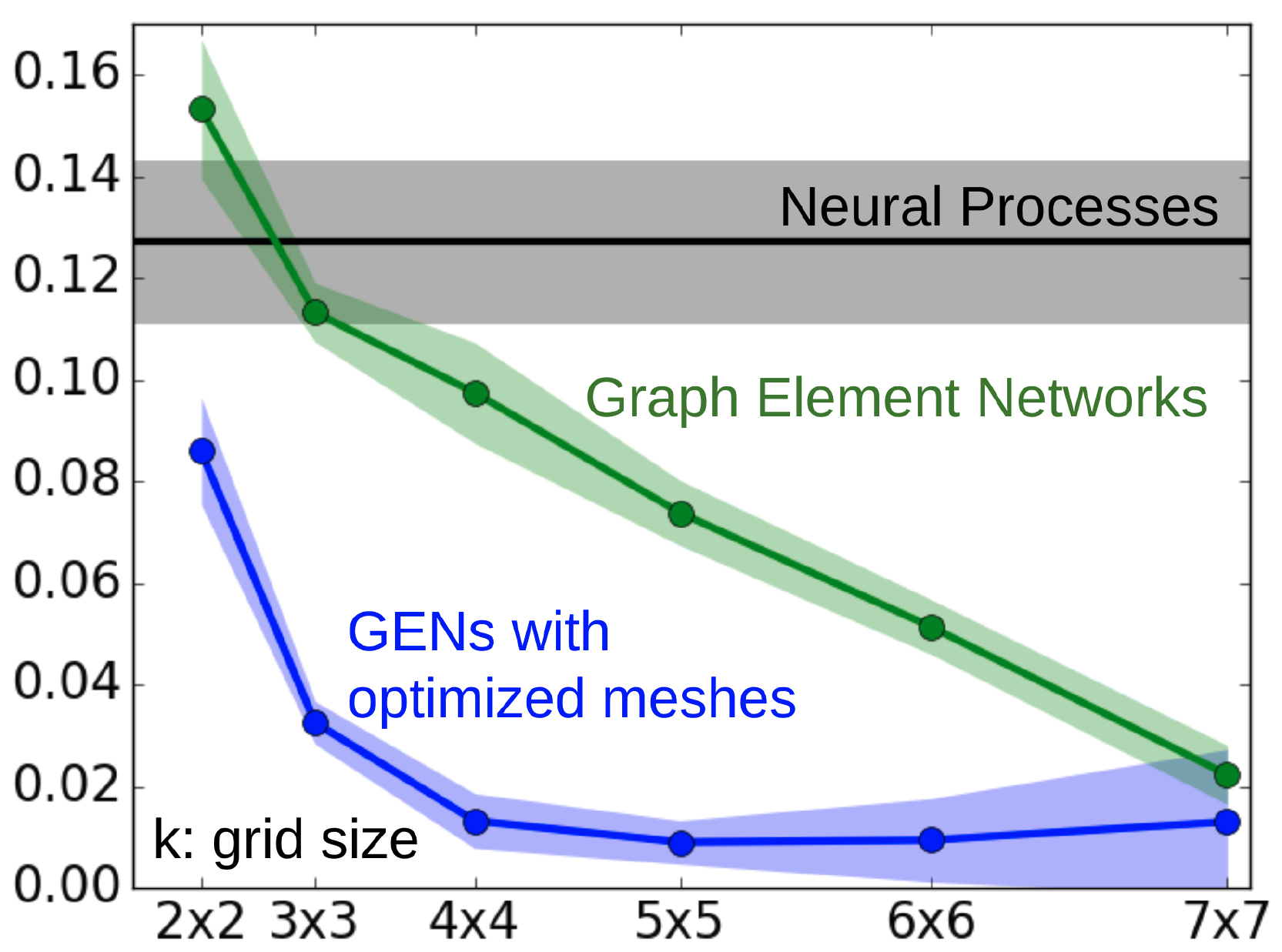}
    \vspace{-0mm}
    \caption{Poisson equation experiments: mean squared error, as a function of the grid size, with one standard deviation in shaded color. Note that GENs use the same set of weights for all mesh sizes. Results for bigger optimized meshes tended to be a bit less stable, resulting in higher variance. }
    \vspace{-0mm}
    \label{fig:Poisson_graphs}
\end{figure}

To apply a GEN to this problem, we must encode $\psi$ and the boundary conditions in the input functions $\vf$ and then train it to generate an output function $\vg$ that approximates $\phi$.  The underlying space is $\sX = [0, 1]^2$.  
We use two input spaces: $\sI_1$ encodes $\psi$, so that an input element $((x, y), 1, (\mu, 0, 0))$ means that $\psi(x, y) = \mu$;  $\sI_2$ encodes the boundary conditions, so that an input element $((x, y), 2, (0, \omega, 1))$ means that $\phi(x, y) = \omega$. The single output space $\sO = \mathbb R$ represents the temperature at any point in $\sX$. We use a set of $k^2$ nodes placed on a uniformly-spaced grid, and a number of message passing steps equal to the diameter of the graph, $T = 2(k-1)$.


We explored two representations. The first places nodes in a regular grid and is a generalization of barycentric coordinates to the rectangular case (details in appendix). The second is a ``soft'' nearest neighbor representation function $r(x, y) = \text{softmax}(D((x, y)))$ where $D((x, y))$ is a vector whose $i-th$ entry is $- \text{dist}((x, y), x_i)$.  The grid-based representation has the advantage of being sparse while the soft nearest-neighbor representation can be applied independently of the placement or topology of the nodes.  Since they did not have significantly different results in our experiments, we only report those with the soft nearest neighbor, which is applicable to all experiments in this section.

\begin{figure}
    \centering
    \includegraphics[width=\linewidth]{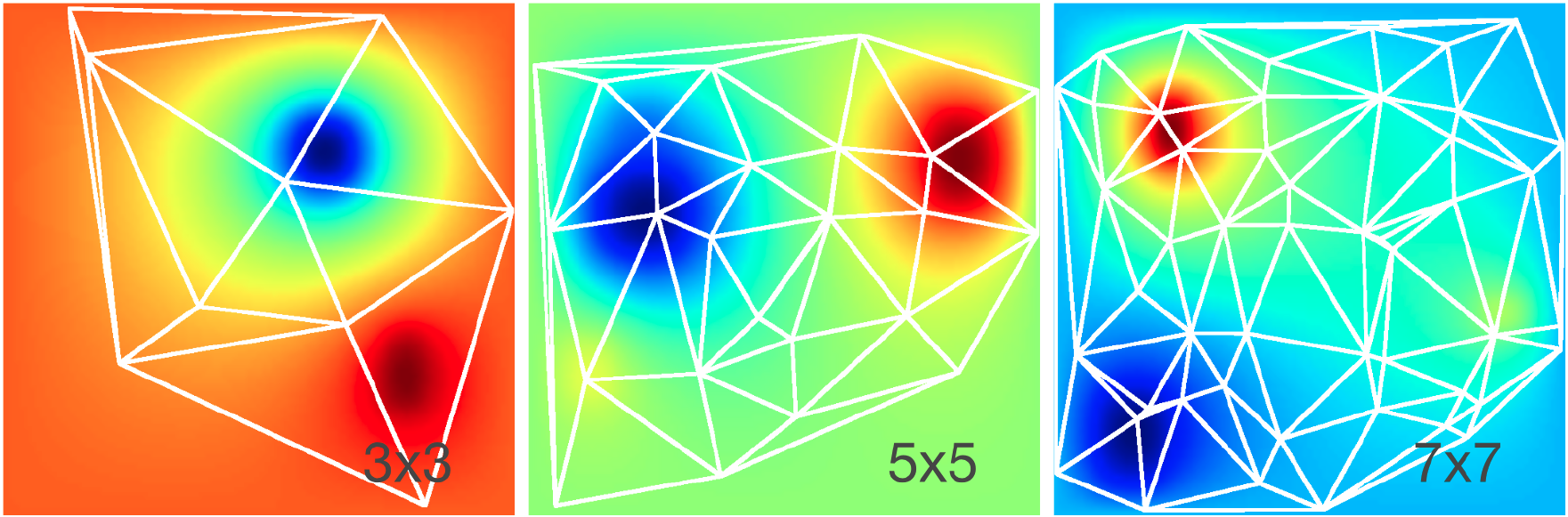}
    \vspace{-0mm}
    \caption{Each white graph shows the placement and connectivity of nodes, after optimization; they move towards more non-linear parts of the space. }
    \vspace{-0mm}
    \label{fig:optimized_meshes}
\end{figure}

The encoder, decoder, node, and edge neural networks are trained on a set of houses (all squares, but with varying number and placement of heaters) and evaluated on a different set. For each house we have a set of 32 scenarios, which share the same heater placements, but vary the temperature of the heaters and external walls (same house at different times of year).  Figure~\ref{fig:Poisson_graphs} shows results for GENs with varying grid density $k$ in comparison with the baseline method.  
We observe that GENs with only a 4x4 mesh already improve upon the unstructured baseline and continue to improve with density. Note that denser meshes represent the problem instance in a more fine-grained way and involve more computation, but the weights used in each model are the same because the GNN modules are shared. We informally tested on meshes bigger than those seen at training time, but performance degraded, as shown in the appendix; this is an interesting direction for future work.
The GEN model has learned to interpret the input specification of the problem and is able to solve new settings of the associated PDE more than 250 times faster than the high-resolution solver.


\begin{figure}
    \centering
    \includegraphics[width=\linewidth]{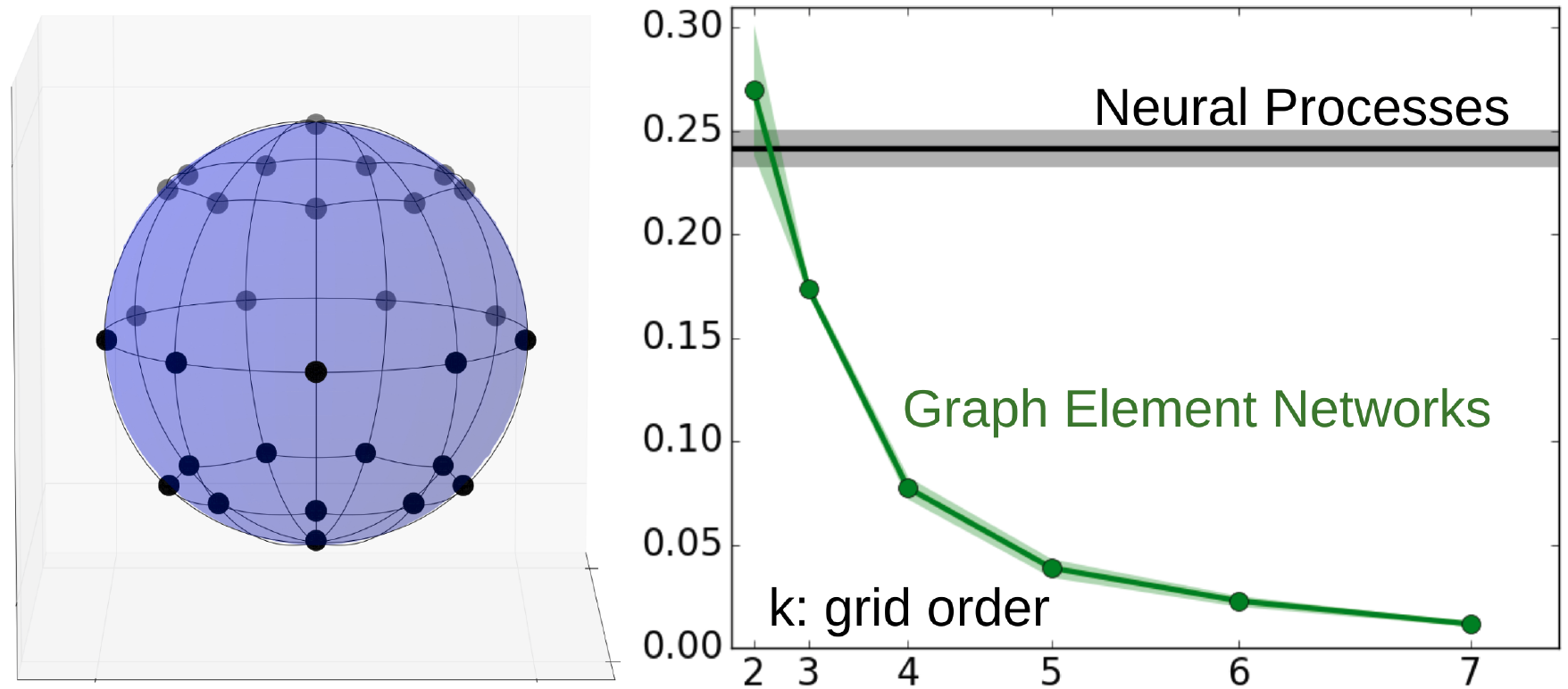}
    \vspace{-0mm}
    \caption{Left: grid of order 5, right: mean squared error, with one standard deviation in shaded color}
    \vspace{-0mm}
    \label{fig:res_sphere}
\end{figure}

\paragraph{Optimizing node positions} In a second experiment, following section \ref{subsec:optimize}, we optimize the positions of the nodes of each house via gradient descent on data from the same house, projecting them back to the square if the gradient step moves them outside. Note that our dataset consists of a set of houses, each with a set of scenarios (different parameters to heaters and external temperatures, with the positions of the heaters held constant). Therefore, similar to the meta-learning setting, we learn the graph network parameters ("physics") with all houses, optimize node positions with a few scenarios from an unseen house and test on other scenarios from the same house. By optimizing the nodes we are able to place greater emphasis harder parts of the space, customizing the mesh to the task, as seen in figure \ref{fig:optimized_meshes}.

\paragraph{Solving a PDE on a sphere} Finally, we also learn a version of the problem on a sphere, detailed in appendix. Since spheres are compact spaces, there are no boundary conditions and thus datasets differ only in the position and temperature of heaters and coolers.
The model requires very few changes to handle this new space: node positions are a $2k\times k$ grid in  polar coordinates, removing duplicates in Euclidean space, and we connect all pairs of nodes below a threshold distance of $\pi/(k-1)$.
In addition, the softmax representation function remains sensible in this space. Results are shown in figure \ref{fig:res_sphere}.

\subsection{Modeling a non-PDE physical system} \label{subsec:Omnipush_exp}

Graph neural networks have been used to model object dynamics~\cite{chang2016compositional,battaglia2016interaction}, however they are typically restricted to objects of a single or few types seen at training time, as they represent each object with a single node. Inferring a custom dynamics model of an object just from a picture of it would be a useful generalization for robotic applications.

Some recent approaches \cite{mrowca2018flexible,li2018learning} assume objects are made of fixed-size particles and model them with a node for each particle. This requires supervision for each particle, limiting training to simulated environments. By contrast, GENs provide an easy way to model dynamics for objects of arbitrary shapes only from a picture-like description. Moreover, at test time, GENs can trade off computation time against accuracy by using smaller meshes when we do not require high precision or in easy scenarios such as when objects are far from being in collision; see~\citet{ullman2017mind} for a motivation from cognitive science.

More concretely, in this section, we show how to use GENs to learn to predict the effects of a robot pushing objects with varying shape and weight, given a simplified top-down image of the object.  
We use the \textit{Omnipush} data set~\cite{omnipush}, a challenging dataset consisting of noisy, real-world data, containing 250 examples for each of 250 objects.
The robot pushes the objects from a random position for 5cm along a random direction, and the initial and final poses of the object are recorded. 

\begin{figure}
    \centering
    \includegraphics[width=.95\linewidth]{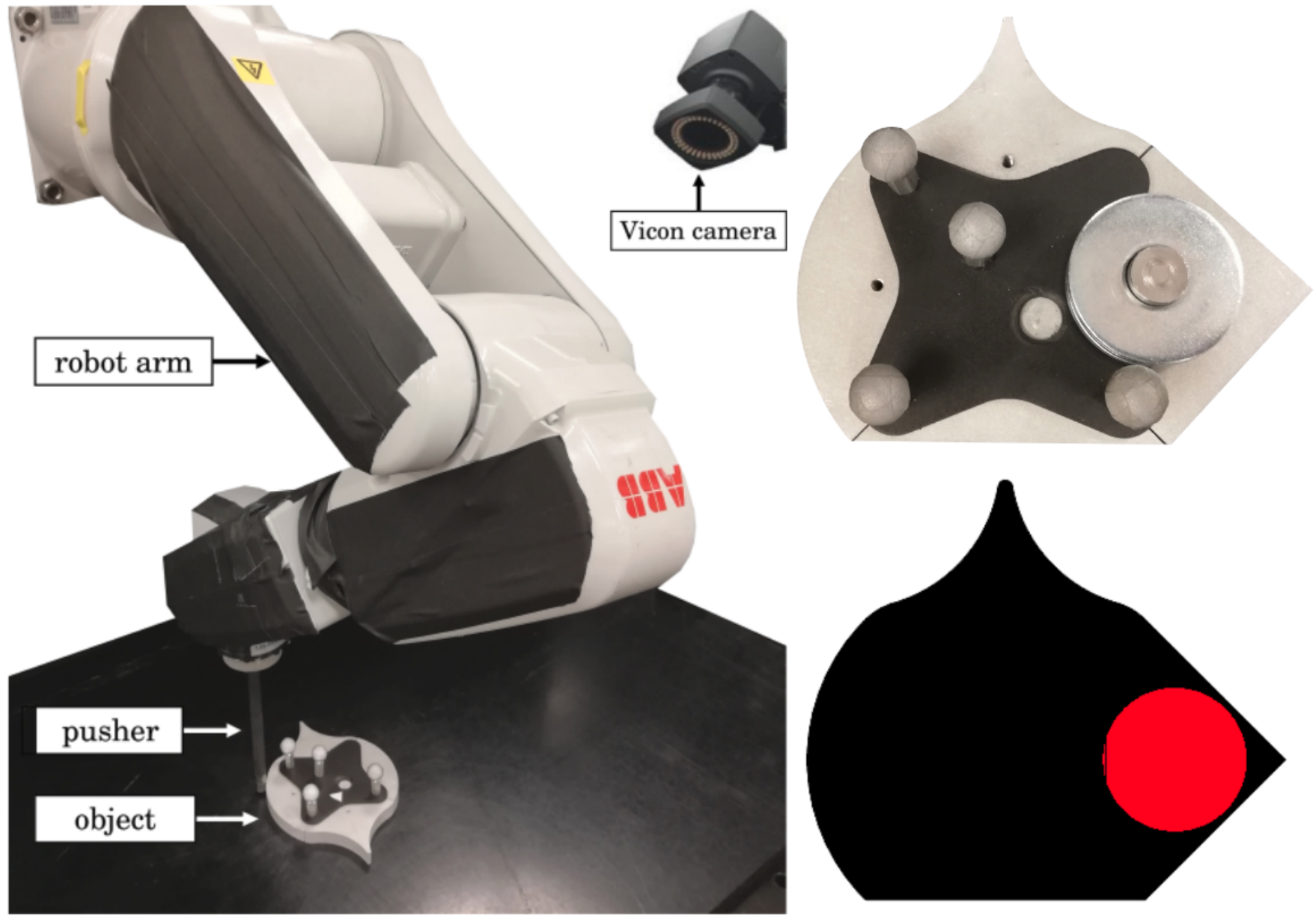}
    \caption{left: \textit{Omnipush} setup, composed of images from \citet{omnipush}. Top right: actual top down view, bottom right: encoded view passed to our model}
    \vspace{-0mm}
    \label{fig:omnipush}
    \vspace{-0mm}
\end{figure}

Because objects have weights attached to them that affect their dynamics (figure \ref{fig:omnipush}), we observe each object from a top down view, where each pixel can be characterized as containing nothing, an unweighted part of the object, part of a light weight, or part of a heavy weight.  The input consists of one-hot encodings of these four pixel classes sampled at 64 locations and an input describing the push with the initial position of the pusher $(x_p,y_p)$ and displacement $(\Delta x_p,\Delta y_p)$. The target consists of $(\Delta x,\Delta y, \Delta \theta)$, the change in the global pose of the object.

The GEN architecture is similar to that of the previous section, except that the output is not a function over $\sX$ space, but a single prediction about the movement of the object.  Thus, we integrate the solution over the space by summing the hidden states of all the nodes prior to feeding it to the decoder.  We report results for a regular grid of size $k=2$, because we found no significant benefit from increasing $k$, possibly because the prediction accuracy is close to the underlying noise of the system as described  by~\citet{omnipush}.

The mean squared prediction error in the object's final position, measured on held-out data, is 0.062 for the modular meta-learning method of~\citet{alet2018modular},  0.078 $\pm$ 0.006 for the NP baseline, and 0.045 $\pm$ 0.002 for the GEN model. 
We observe that the predictions generated by the GEN are significantly more accurate than those of the other two methods.
Note that for the modular meta-learning results reported here, rather than being given images of the objects, the algorithm was given training data segregated into batches of 50 based on the ground-truth object shape and weight.

\subsection{Scene representation with GENs}\label{subsec:GQN_exp}


In this section, we illustrate the ability of GENs to perform a difficult spatial prediction problem and to adaptively grow with the spatial extent of new problem instances, using the 
problem of \textit{neural scene representation}~\cite{eslami2018neural}. The problem consists of receiving $K$ 2D images from different 4D camera poses in a single 3D scene, a maze in our case, then predicting the image that would be seen from a new camera pose. 
The original work on this problem proposed {\it generative query networks} (GQN), which first use a neural process to encode the input views ((pose, image) pairs) and then feed the encoding to a DRAW decoder~\cite{gregor2015draw} that renders the predicted image. For these experiments, we adapted the code of~\citet{Taniguchi18}.


Unfortunately, the DRAW decoder has significant computational requirements.
To reduce the computation required, we simplified the benchmark to require the system to identify the correct image among a set of candidates, rather than to generate a completely new image, similar to contrastive learning methods~\cite{sermanet2017time,oord2018representation}.
More concretely, after it has been trained, the GEN is given as input:  8 training pairs consisting of an image and a camera pose gathered from different viewpoints of a single maze, 576 candidate images from that maze and other mazes, and a query camera pose.  It must predict which of the candidate images corresponds to the query pose.
Each batch of training data consists of data from 64 mazes, with 8 (image, pose) pairs and 1 query pose per maze.  The set of candidate images includes all target images and 8 other randomly sampled images from each maze in the batch. 

Our GEN model treats the $(x, y)$ coordinates of the camera pose as the space $\sX$ and the image and camera roll and pitch angles as the input space $\sI$.  Because we have an image classification task, it is possible to interpret the output space $\sO$ as an output distribution over the candidate images.   This is not strictly correct, however; 
for these and other experimental details, see the appendix.
We first use a CNN to map input images and camera poses into the 256-dimensional GEN latent space $\sL$. 
Given a query camera pose, the latent representation $z$ is determined based on the $(x, y)$ coordinates.  This $z\in\sL$ is concatenated with the camera pose and decoded to an embedding space $\sZ'$.  The candidate images are also mapped into $\sZ'$, allowing us to compute a vector of squared Euclidean distances 
and feed it through a softmax to generate the final prediction.

\begin{figure}
    \centering
    \includegraphics[width=\linewidth]{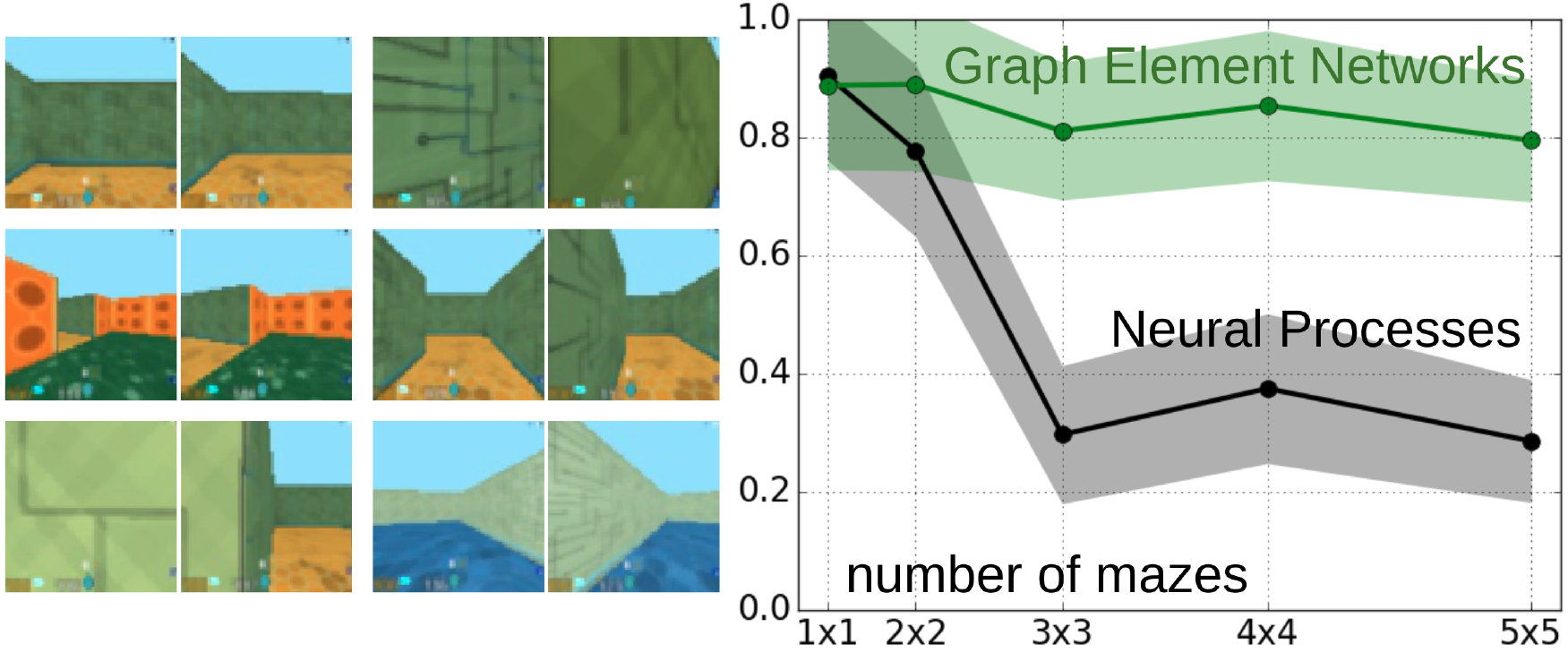}
    \vspace{-2mm}
    \caption{(left) Randomly selected misclassified examples by the GEN model: ground truth and prediction respectively. (right) Accuracy as a function of supermaze size.  The performance of the baseline decreases with number of mazes while that of GENs is roughly constant.
    }
    \label{fig:GEN_mistakes}
    \label{fig:hotels}
    \vspace{-2mm}
\end{figure}

In our implementation, we use a $5 \times 5$ grid of nodes, uniformly spaced in camera $(x, y)$ space and connected as a grid, and 9 steps of message passing.
GENs achieved a statistically equivalent classification accuracy to NP's 
in about half the number of epochs, as shown in the appendix.  This is a difficult problem, because some images are very similar and often 8 views are not enough to understand a maze well (the original paper used between 1 and 20). Figure~\ref{fig:GEN_mistakes} illustrates several randomly-selected classification errors.

One strength of the GEN model is the flexibility to model spaces of different shapes and sizes than it has been trained on, whereas most deep learning models (such as NPs) have a fixed capacity and thus cannot generalize over problem shape and size.
We studied this aspect of GENs by creating ``supermazes'':  concatenations of $m \times m$ of the original mazes, arranged in a grid.  We trained on example $1 \times 1$ and $2 \times 2$ supermazes, but tested on examples with $m = 1 .. 5$.  We still trained the system with $8$ views and one query per maze.
To avoid testing camera positions that were outside the range of training positions, all training and testing supermazes were randomly positioned within a $5x5$ grid. In figure~\ref{fig:hotels} we can see that the performance of the NP model degrades as $m$ increases, but the performance of the GEN model does not.

\section{Conclusion}
We have proposed a new method for leveraging the combinatorial generalization and relational inductive biases of graph neural networks, without the requirement of matching {\em a priori} entities to nodes. This allows us to create models with structured, flexible computation. Structured because the connectivity of the graphs along with the reuse of neural modules across nodes and edges provides a useful inductive bias for problems with a spatial component. Flexible because, since nodes are not tied to entities, we can optimize the graphs to obtain higher accuracy or adapt the amount of computation to our desired accuracy.


\section*{Acknowledgements}
We thank Kelsey Allen, Xavier Fern\'andez-Real, Clement Gehring, Peter Karkus, Ari Morcos and Tom Silver for useful discussions and Eduardo Delgado and Albert Villalobos for their help setting up the experiments.

We gratefully acknowledge support from NSF grants 1523767 and 1723381; from AFOSR grant FA9550-17-1-0165; from ONR grant N00014-18-1-2847; from Honda Research; and from the MIT-Sensetime Alliance on AI. FA acknowledges support from a "laCaixa" fellowship (ID 100010434, code LCF/BQ/AN15/10380016). MB is supported by the NSF award
[IIS-1637753] through the National Robotics Initiative. Any opinions, findings, and conclusions or recommendations expressed in this material are those of the authors and do not necessarily reflect the views of our sponsors.

\newpage
\bibliography{main}
\bibliographystyle{icml2019}
\appendix
\newpage
\section{Experimental details}
Code can be found at \url{https://github.com/FerranAlet/graph_element_networks}.
\subsection{Details common to all experiments} \label{subsec:generic_exp_app}
Although the experiments are pretty diverse, most of the code and decisions are shared between them. We first talk about the details common to all experiments.

We explored two representation functions. The first assumes that nodes are placed in a regular grid and uses the latent values at the four nodes of the grid cell containing a point to compute its latent value.  So the representation function $r(x, y) = v$;  letting $(i_1, i_2, i_3, i_4)$ be the node indices of the grid cell containing $(x, y)$, and letting $(x_{i_j}, y_{i_j})$ be the location of node $i_j$, then $v_{i_j} = \lvert x - x_{i_j} \rvert \cdot \lvert y - y_{i_j} \rvert$ and $v_\ell = 0$ for all other indices $\ell$.   This is a generalization of barycentric coordinates to the rectangular case.  
The second is a ``soft'' nearest neighbor representation function $r(x, y) = \text{softmax}(D((x, y)))$ where $D((x, y))$ is a vector whose $i-th$ entry is $- \text{dist}((x, y), n_i)$.  The grid-based representation has the advantage of being sparse while the soft nearest-neighbor representation can be applied independently of the placement or topology of the nodes.  Since they did not have significantly different results in our experiments, we decided to use the ``soft'' nearest neighbor in all experiments of the Poisson experiments, since it also works for non-grid meshes and used the barycentric representation for the pushing and scene experiments, to show it can also work with sparse representations.

For our implementation we used PyTorch \cite{Paszke2017AutomaticDI}, we used ReLU non-linearities, trained with Adam \cite{Kingma2014AdamAM} with learning rate 3e-3. 

We tried both training independent networks for each mesh and training a single model for all meshes, with no significant difference between both. We only put the results of a single model, since it reinforces our message of a single model being able to trade off computation and accuracy. We note that, as the density of the mesh increases, the number of computations and depth for GENs increases, but since we are reusing the same weights, the number of weights is always smaller than those of the baseline. The depth of the baseline is similar to a GEN of 4x4 nodes and we found that adding extra depth did not help much.
\subsection{Poisson experiments on a square}
The GNN within the GEN is specified by:
    latent space $\sL = \mathbb R^{32}$;
    a set of $k^2$ nodes placed on a uniformly-spaced grid;
    edges generated by the Delaunay triangulation of the nodes;
    encoders $e_1$, $e_2$ and decoder $d_1$ are two-layer neural networks (with a hidden state of 48 and 32 units respectively);
   the GNN message size is 16, and both edge and node modules have one hidden layer of size 48 and 64 respectively; and the diameter of the graph is  $T = 2(k-1)$.

    The test set results were averaged over 256 sample points per scenario, for 32 scenarios of 50 houses.
    We trained the baseline for 3000 epochs and GENs for 3000/6=500 epochs (since we trained on 6 mesh configurations, from 2x2 to 7x7).
    
\subsection{Poisson experiments on a sphere}\label{subsec:sphere_app}

For the spheres, we generated our own dataset of pairs of Laplacians and PDE solutions. Similar to the square house experiments, we had 250 houses with 32 scenarios each; trained on 200 and tested on 50. We had 128 inputs and 128 queries. In contrast to the square experiments, since we did not have easy access to a PDE solver for a sphere, we instead specified the solution of the PDE and numerically computed the Laplacian. The solution of the PDE was of the form:
$$f(x) = \sum_{i=1}^8 k_{i,s}\cdot (\vx\cdot \vv_i)^3$$
with $k_i$ a random gaussian scalar, which also varied with scenario $s$ and vector directions $\vv_i$ which played the role of heater/cooler positions. Moreover, since the dot product on a sphere is symmetric and we raise it to the cube (an odd function), we know:
$$\int_{x\in \mathcal{S}^2} \left[(x\cdot \vv_i)^3\right]=0$$
this in turn forces:
$$\int_{x\in \mathcal{S}^2} f(x)= \sum_{i=1}^8 k_{i,s} \int_{x\in \mathcal{S}^2}\left[(x\cdot \vv_i)^3\right]= \sum_{i=1}^8 0 = 0$$
This is important because since the sphere has no boundary conditions, solutions to the Poisson equation are only defined up to a constant. We thus define that the solution is the \textit{unique} function which satisfies the equation \textit{and} has integral 0.

Computing the Laplacian on a sphere is not trivial to do analytically since the partial derivatives have to be taken on the sphere, not on Cartesian space. Since the Laplacian is related to the difference between $f(p)$ and the mean value of $f(x)$ in the epsilon ball around $p$, another interpretation is to say that the relevant neighbors are only those that lie on the sphere.
We compute the Laplacian numerically by computing the tangent plane at $\vx$ by using Gram-Schmidt to complete an ortonormal basis which includes $\vx$: $\{\vx,\va,\vb\}$. Then, we numerically approximate both second derivatives by taking $\epsilon$ steps in $\va,\vb$, projecting back to the sphere and evaluating there:
\begin{align*}
  \nabla^2 f(\vx) \approx \frac{1}{4\epsilon^2}\big( & 
f(\frac{\vx+\epsilon\va}{|\vx+\epsilon\va|}) + f(\frac{\vx-\epsilon\va}{|\vx-\epsilon\va|}) + \\
& f(\frac{\vx+\epsilon\vb}{|\vx+\epsilon\vb|}) 
+ f(\frac{\vx-\epsilon\vb}{|\vx-\epsilon\vb|}) - 4f(\vx) \big) 
\end{align*}
We used $\epsilon = 3e-5$, but any $\epsilon \in [3e-6,3e-4]$ gave essentially the same results.

To make it as similar as possible to the experiments on a square, node positions also followed a grid in spherical coordinates: 
$$x = \sin \theta \cos \phi$$
$$y = \sin \theta \sin \phi$$
$$z = \cos \theta$$
The mesh of order $k$ $\theta \in \{0,\pi/(k-1),2\pi/(k-1),\dots,\pi \}$ and $\phi\in\{0,\pi/(k-1),2\pi/(k-1),\dots,2\pi \}$ and then removing points that were duplicates in Cartesian space; resulting in $k$ longitudes of $1,2k-1,2k-1,\dots,2k-1,1$ nodes. Note that, for the sphere, the soft nearest neighbor distance function is not Euclidean, but the distance on the sphere surface (the arc-cosine of the dot product, since we are on the unit sphere). We trained both the baseline for 10000 epochs and GENs for and GENs for 10000/6=1666 (6 mesh configurations, with sizes randing from 2 to 7).
\subsection{Optimizing node positions}
We trained the GEN with optimizable positions for 10000/12=833 epochs (we trained with 12 mesh configurations).  The optimization of the node positions used learning rate 3e-4, while the weights still used learning rate 3e-3. This came out of an informal search where we wanted nodes positions to not prematurely converge, yet also be able to go to any possible configuration. Smarter ways to perform this optimization (potentially including non-local optimization) are an interesting avenue for future work.

\begin{figure*}[t]
    \centering
    \includegraphics[width=\textwidth]{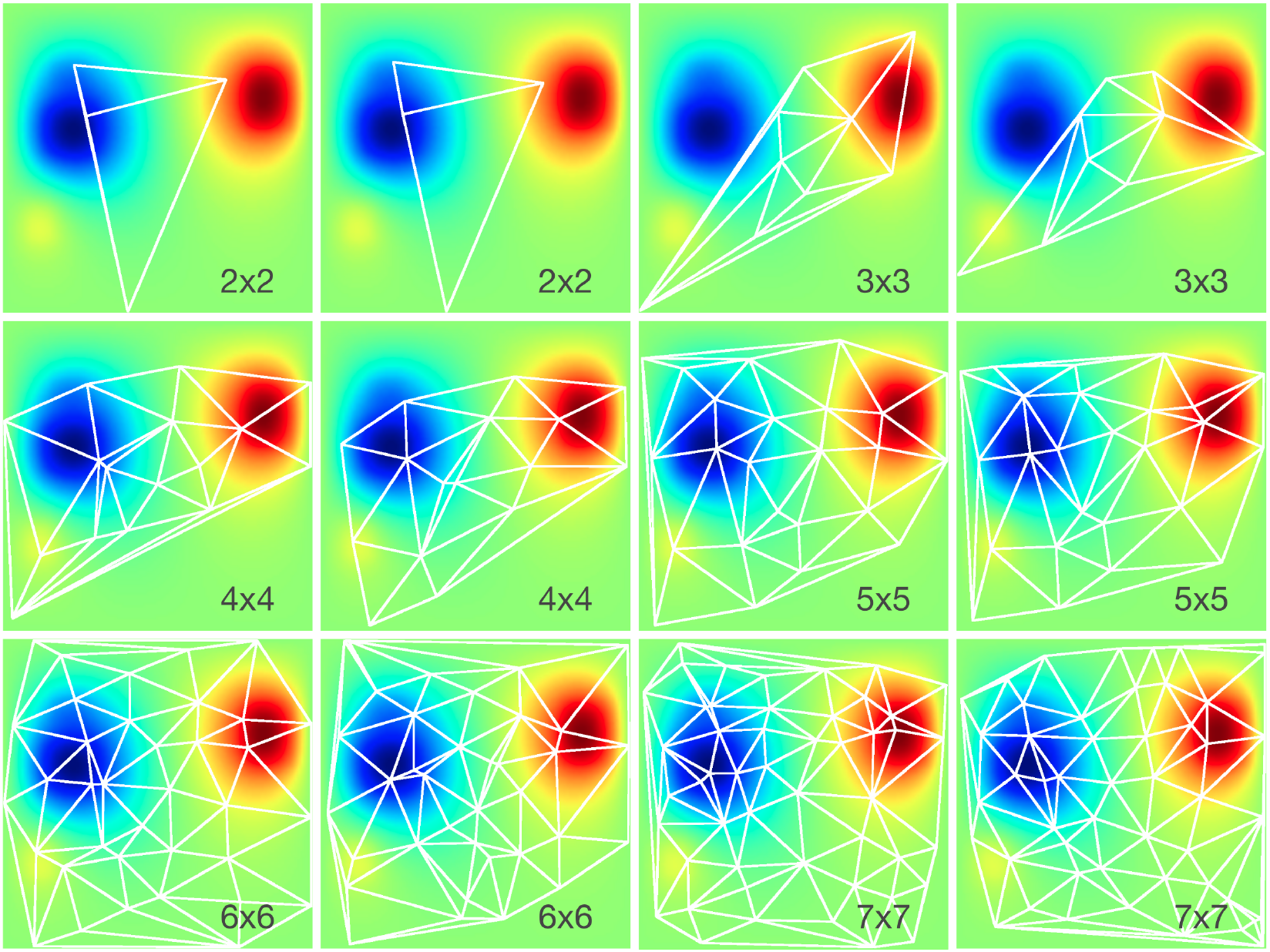}
    \caption{\textbf{One room with all the optimized meshes}. We notice that each size makes the most of its amount of nodes by focusing on the most complex parts of the space. For each size, the pair of meshes resembles each other, but (except for the 2x2) are qualitatively different.}
    \label{fig:one_room_all_meshes_numbered}
\end{figure*}

For each mesh size we initialize two random positions, which tend to converge to similar but different final positions, as seen in figure \ref{fig:one_room_all_meshes_numbered}. The initial positions of the nodes were generated using the ghalton python library~\cite{ghalton}, which uses Generalized Halton Sequences to create points that are random but well spread out along a region. This is because uniformly random points tend to get clustered by pure chance.

The connectivity was the Delaunay triangulation, which has many desirable properties for our purposes: it has low graph diameter~\cite{bose2007stabbing}, it is very stable (small perturbations tend to produce little or no changes) and changes are local (moving a point can only affect edges nearby). Moreover it can be computed efficiently and it is the dual of Voronoi diagrams, which map each point to its nearest neighbor in space, thus being linked to a good representation function. After each gradient step, we recomputed the connectivity. Note that these changes are non-differentiable; we considered smoothing the connectivity by weighting each edge proportionally to the angle it represents in the Voronoi diagram, but performance was great without it and we preferred to keep it simple. Pure back-propagation would move some points outside the $[0,1]^2$ region; therefore, after back-propagation we clamp points back inside the region.

\subsection{Scene representation experiments} \label{subsec:GQN_exp_app}
Our GEN model treats the $(x, y)$ coordinates of the camera pose as the space $\sX$ and the image and camera roll and pitch angles as the input space $\sI$.  Because we have an image classification task, it is possible to interpret the output space $\sO$ as an output distribution over the candidate images.   This is not entirely correct, however, because the set of candidate images varies during training, so this is not precisely a fixed output space.
The encoder of our GEN model is the same as the one used by the original NP:  it uses a CNN to map input images and camera poses into a 256-dimensional latent space.  Once an (image, pose) pair is encoded, it is stored into the latent state of the nodes in the GEN using the representation function.  Given a query camera pose, the latent representation $z$ is determined based on the $(x, y)$ coordinates.  This $z\in\sL$ is concatenated with the camera pose and decoded to an embedding space $\sZ'$.  The candidate images are also mapped into $\sZ'$, allowing us to compute a vector of squared Euclidean distances betweeen the query output and the candidate images in $\sZ'$.  This vector is fed through a softmax to generate the final vector of output values.
For the GEN structure, the edge set is composed of the all the bi-directional edges between pairs of adjacent nodes on the grid described in 4.3. The rectangular generalization of barycentric coordinates, described in A.1 was used to compute the latent value of any query position $(x,y)$. When obtaining the encoding of input images (with their camera pose coordinates) with the GEN, the GNN messages used have 256 dimensions, and both the node and edge neural networks are 2-layer feedforward networks with a hidden layer of 512 units. To compensate for the these GNN computations, the NP capacity is increased by always concatenating 32 copies of input query coordinates to their corresponding embeddings. For the GEN we only provide 8 copies, and we experimentally observed that these copies slightly benefit the GEN and strongly benefit the neural process. We believe that, without copies, the signal from the query pose is drowned by the embedding, which has many more dimensions. We did not notice further benefits beyond 32 copies, which makes sense since $32\cdot 7$ is already of the same order of magnitude as 512. For both the GEN and baseline experiments the encodings are post-processed through feed-forward networks, which have depth 2 and width 256 for the GEN and depth 4 and width 512 for the NP, respectively. 

When running the scene experiments for many epochs, without batch normalization (not added in other experiments) the performance of both GENs and the baseline smoothly increased until $80\%$ accuracy on 1x1, but then weights exploded and performance plummeted for both models. Batch normalization solved this issue for both models.
\begin{figure}
    \centering
    \includegraphics[width=.9\linewidth]{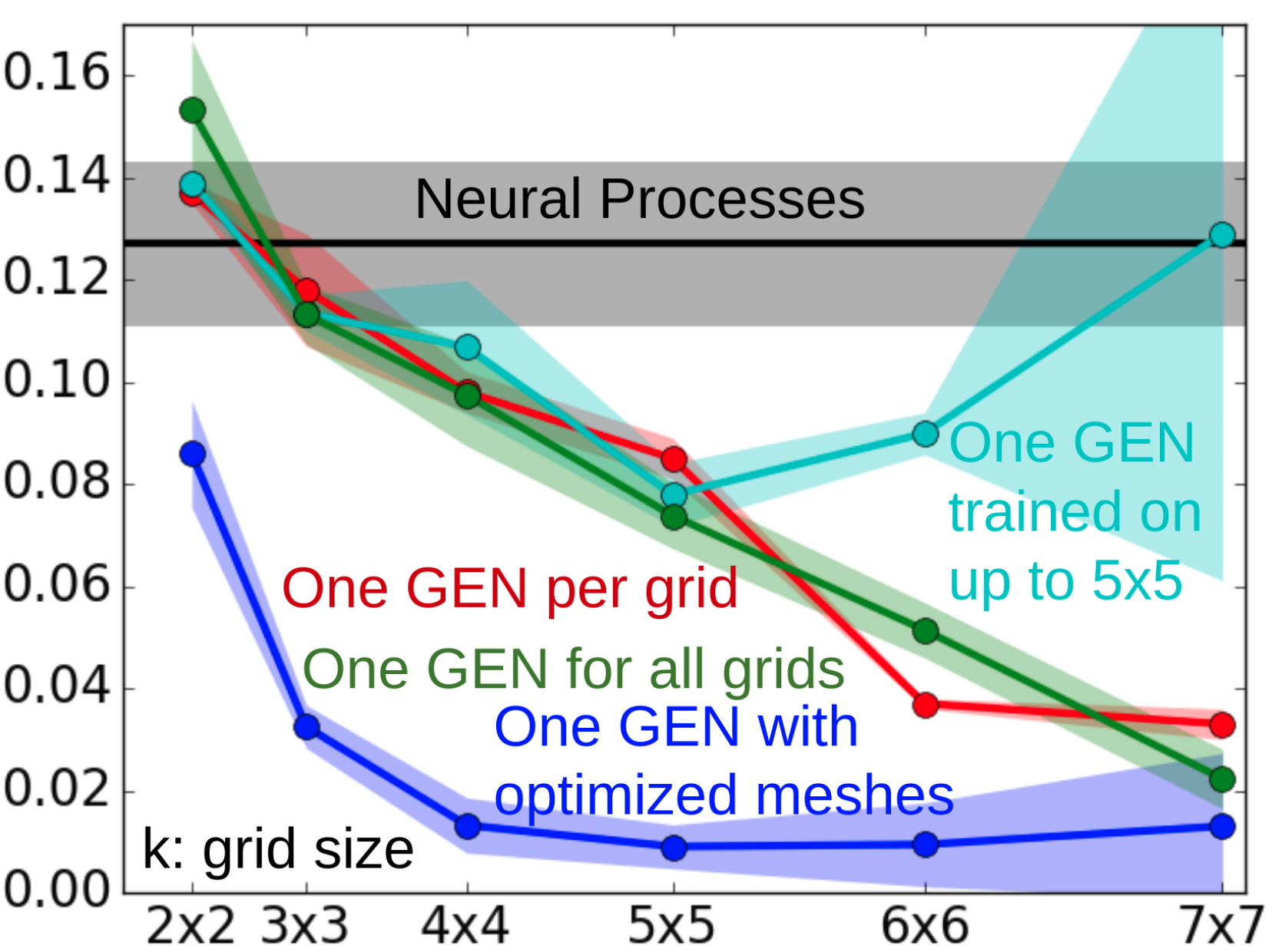}
    \caption{Comparison betweeen original GENs and two other reasonable approaches. We see that a GEN using the same weights for all mesh sizes (green) can do as well as using a custom GEN for each mesh size. However, the same weights do not generalize to much bigger meshes (cyan, trained only on 2x2..5x5). Contrary to other GNN-based systems that generalize to bigger input sizes than those seen at test time, we do increase the number of propagation steps and graph size increases quadratically; both things can make propagation unstable. This is an interesting point to address in future work.}
    \label{fig:poisson_dirty_laundry}
\end{figure}

\begin{figure*}[t]
    \centering
    \includegraphics[keepaspectratio=true, scale = 0.6]{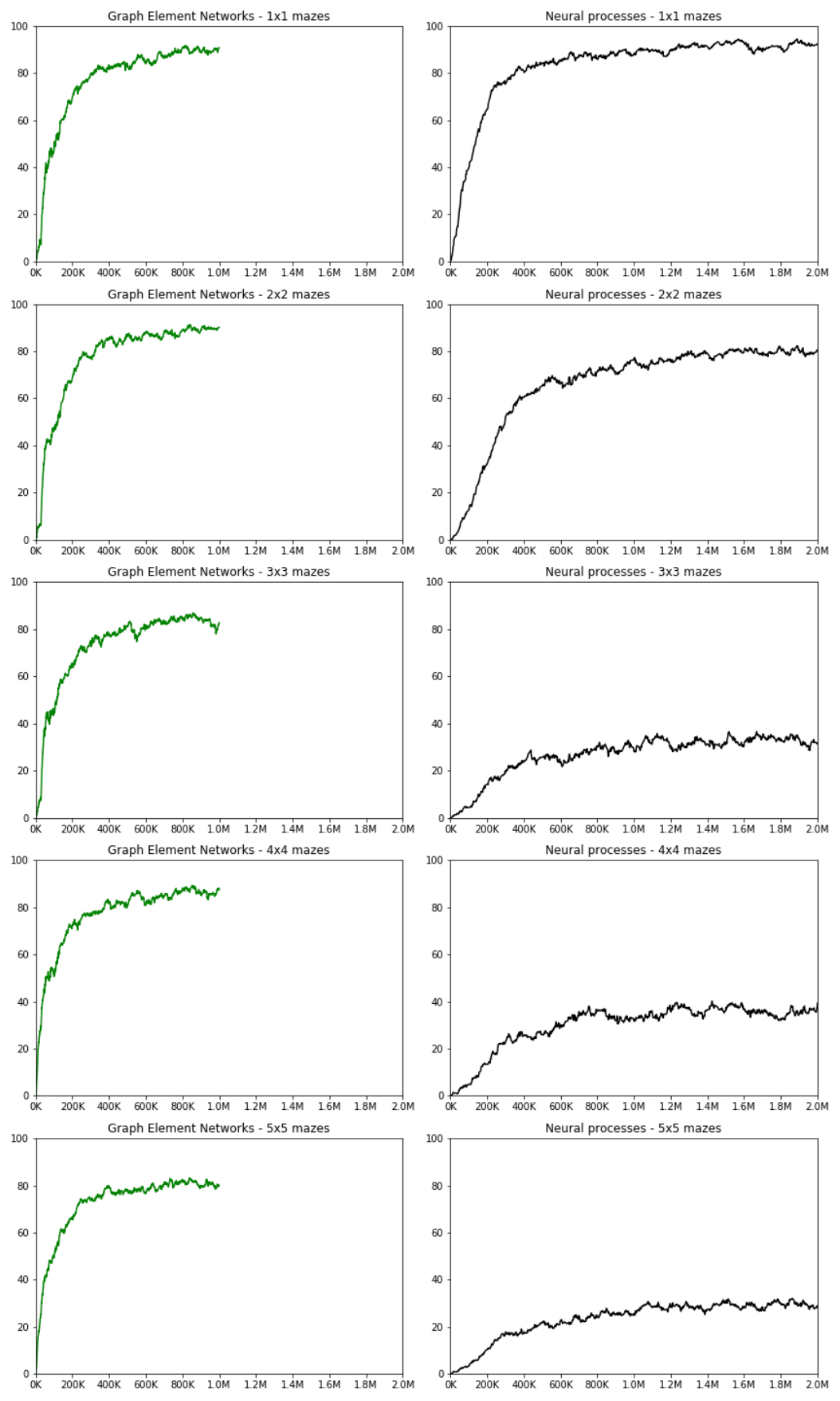}
    \caption{Test accuracy plotted against number of mini-batch gradient descent steps, for Graph Element Networks experiments (left) and the baseline Neural Processes experiments (right). The graphs from top to bottom show performance on supermazes of increasing sizes (1x1 to 5x5).}
    \label{fig:graphs_accuracies_scene_classification}
\end{figure*}
\end{document}